\documentclass[11pt]{article}

\usepackage[utf8]{inputenc}
\usepackage[T1]{fontenc}
\usepackage[margin=1in]{geometry}
\usepackage{amsmath,amssymb}
\usepackage{booktabs}
\usepackage{multirow}
\usepackage{graphicx}
\usepackage{xcolor}
\usepackage{url}
\usepackage[numbers,sort&compress]{natbib}
\usepackage{hyperref}
\usepackage{microtype}
\usepackage{array}
\usepackage{tabularx}

\hypersetup{
  colorlinks=true,
  linkcolor=blue,
  citecolor=blue,
  urlcolor=blue
}

\title{Routing Ceilings Are Domain-Independent:\\
Structural Prior Injection in Code Security\\
Vulnerability Detection}

\author{
  Manuel Israel C\'{a}zares \\
  Bytepro AI \\
  Mazatl\'{a}n, Sinaloa, Mexico \\
  \texttt{hello@bytepro.ai}
}

\begin{document}

\maketitle

\begin{abstract}
Large language models (LLMs) exhibit a well-documented gap between latent
capability and consistent activation.
C\'azares~\citep{sair2026} proposes the \emph{router hypothesis}: that models
\emph{route problems to cached structural patterns rather than deriving
answers compositionally}.
That work reports that injecting structural priors via cheatsheets raises
performance dramatically on in-distribution data, yet collapses even below
the zero-shot baseline when evaluated on an out-of-distribution (OOD) test
set---and that a more heavily engineered cheatsheet (AN45c) that peaks
locally underperforms a simpler predecessor (AN38) under official
distribution shift.

We ask whether this phenomenon is \emph{cross-domain}.
We reproduce the experimental design of~\citep{sair2026} in the domain of
\emph{source-code security vulnerability detection}, evaluating three
frontier LLMs (GPT-OSS-120B, Llama-3.3-70B, Gemma-4-31B) across three
vulnerability categories (two CWEs --- CWE-798, CWE-284 --- plus one
non-CWE anti-pattern, N+1) spanning syntactic, contextual, and semantic
vulnerability types on a synthetic corpus of 348 labeled pairs
(278 training, 70 held-out evaluation), and then
transferring the cheatsheet-augmented prompts to real-world
CVE (Common Vulnerabilities and Exposures) data
from the VUDENC benchmark (CWE-89, CWE-22).

Our main findings replicate and extend those of~\citep{sair2026}:
\textbf{(F1)} Structural priors reduce routing failures on synthetic data,
lifting the semantic-vulnerability recall from 20.0\,\% to 100.0\,\% across all
three models.
\textbf{(F2)} Zero-shot performance degrades along a semantic
complexity gradient (syntactic $>$ contextual $>$ semantic)
for GPT-OSS-120B, and partially for Gemma-4-31B;
Llama-3.3-70B shows an inverted pattern on N+1.
\textbf{(F3)} The same cheatsheets that saturate synthetic performance
\emph{amplify} distribution-shift collapse on real CVE data: for CWE-89,
GPT-OSS-120B drops from 100\,\% synthetic F1 to 48.9\,\% on VUDENC
($-$51.1\,pp), compared with 61.5\,\% zero-shot ($-$29.4\,pp).
\textbf{(F5)} Iterative recalibration of the routing priors using real
error patterns produces a v2 cheatsheet that performs \emph{worse} than v1
on the same real distribution (41.7\,\% vs 48.9\,\% vulnerable-class F1 for
CWE-89 with GPT-OSS-120B), mirroring the AN45c underperformance relative to
AN38 reported by~\citep{sair2026}.

Together these results provide evidence that the \emph{cross-distribution
trade-off surface} documented by~\citep{sair2026} generalizes to code
security, and that the router hypothesis is cross-domain.
We discuss the implications for prompt-engineering-based detection pipelines
and argue that the structural nature of the collapse motivates
distribution-aware training rather than prompt calibration as a more promising remedy.
Code, cheatsheets (v1/v2), and evaluation scripts are available at
\url{https://github.com/bytepro-ai/bitcoder-v2-research}.
\end{abstract}

\section{Introduction}
\label{sec:intro}

The ability of large language models (LLMs) to solve complex reasoning
tasks has grown substantially, yet a persistent gap remains between
\emph{peak} performance---achievable under carefully crafted prompts---and
\emph{robust} performance across distributional variation.
Understanding the nature of this gap is central to deploying LLMs reliably
in high-stakes domains such as automated software security analysis.

\subsection{The Router Hypothesis}

C\'azares~\citep{sair2026} introduced the \emph{router hypothesis} in the
context of the SAIR Equational Theories Stage~1 competition: language models
\emph{route problems to cached structural patterns rather than deriving
answers compositionally}.
Injecting a \emph{structural prior}---a compact cheatsheet enumerating
canonical solution routes---can dramatically raise in-distribution (ID)
performance.
However, the same cheatsheet creates a \emph{distribution-shift hazard}: when
the test distribution shifts even modestly, the injected routing prior
misaligns with the new patterns, systematically directing the model toward
incorrect solution paths.
Prior work further found that a more heavily engineered variant (AN45c),
which peaks on a local hard split, performs \emph{worse} than its simpler
predecessor (AN38) on the official evaluation set, suggesting that the
collapse reflects a structural property rather than a calibration artifact.

\subsection{Extending the Hypothesis to Code Security}

Software vulnerability detection is a natural and practically important
stress test for the router hypothesis.
Like formal mathematics, it requires precise pattern-matching against
structured artifacts (source code), hierarchical categorization (CWE
taxonomies), and reasoning under distributional shift---since vulnerabilities
in production code look substantially different from textbook examples or
synthetically generated code snippets.

Prior LLM-based vulnerability detection work has demonstrated strong
performance on curated benchmarks~\citep{thaller2019code,wartschinski2022vudenc},
but systematic analysis of the
cheatsheet-induced trade-off between ID and OOD performance has not been
undertaken.
We fill this gap.

\subsection{Contributions}

This paper makes the following contributions:

\begin{enumerate}
  \item We replicate the experimental protocol of~\citep{sair2026} in the
        code security domain, establishing a \emph{cross-domain baseline}
        for the router hypothesis (\S\ref{sec:methodology}).

  \item We demonstrate that structural priors saturate synthetic vulnerability
        detection (up to 100\,\% F1) across three vulnerability categories
        (two CWEs --- CWE-798, CWE-284 --- plus one non-CWE anti-pattern,
        N+1) spanning syntactic, contextual, and semantic complexity,
        providing evidence for the router hypothesis at scale
        (\S\ref{sec:results:synthetic}).

  \item We characterize the \emph{semantic complexity gradient}: zero-shot
        LLM performance on vulnerability detection degrades predictably for
        GPT-OSS-120B and partially for Gemma-4-31B as a function of the
        semantic distance between code pattern and vulnerability class
        (\S\ref{sec:gradient}).

  \item We document \emph{amplified distribution-shift collapse}: the same
        structural priors that saturate synthetic performance drive real-world
        F1 below the zero-shot baseline by up to 58\,pp, exceeding the
        magnitude of the analogous collapse reported by~\citep{sair2026}
        (\S\ref{sec:results:real}).

  \item We show that \emph{iterative recalibration amplifies rather than
        corrects} the collapse, with the v2 cheatsheet calibrated against real
        error patterns underperforming v1 on the same real distribution---an
        exact structural parallel to the AN45c finding of~\citep{sair2026}
        (\S\ref{sec:analysis}).

  \item We characterize zero-shot misrouting as a genuine but correctable
        failure mode: models frequently detect real vulnerabilities but
        assign incorrect CWE labels under zero-shot conditions
        (13 synthetic and 19 real-world events), a pattern structural
        priors eliminate entirely---though we note this correction is
        partly attributable to the prior directly supplying the category
        label (\S\ref{sec:misrouting-note}).
\end{enumerate}

\subsection{Significance}

We refer to this evaluation framework---spanning datasets,
cheatsheets, and cross-distribution evaluation protocol---as
\emph{BitCoder}.
The practical implication is stark.
Prompt-engineering-based security pipelines---checklists, few-shot exemplars,
structured routing prompts---may achieve impressive numbers on synthetic or
held-out benchmarks while silently degrading on the real CVE distributions
they are ultimately deployed against.
Our results suggest this degradation is not a tuning problem solvable by
better prompt engineering, but a structural consequence of distributional
mismatch between the routing priors and the target domain.
This motivates an alternative path: distribution-aware training on hybrid
synthetic+real datasets, which we discuss in \S\ref{sec:discussion}.

More broadly, by reproducing the cross-distribution trade-off surface of
~\citep{sair2026} in an entirely different domain (mathematics $\to$ code
security), with different models, datasets, and vulnerability taxonomies,
we provide evidence for the cross-domain replication of the router
hypothesis---suggesting it reflects a fundamental property of how LLMs
activate and apply learned sub-skills rather than an artifact of any
particular task or architecture.

\subsection{Paper Organization}

The remainder of the paper is structured as follows.
\S\ref{sec:background} reviews the router hypothesis and the prior study
of~\citep{sair2026}.
\S\ref{sec:methodology} describes our datasets, models, and experimental
conditions.
\S\ref{sec:results:synthetic} presents results on synthetic
(in-distribution) data.
\S\ref{sec:results:real} presents results on real-world (VUDENC) data.
\S\ref{sec:analysis} provides the v1\,vs.\,v2 comparison and error
analysis; zero-shot misrouting is analyzed in \S\ref{sec:misrouting}.
\S\ref{sec:discussion} discusses implications and limitations.
\S\ref{sec:conclusion} concludes.

\section{Background}
\label{sec:background}

\subsection{The Router Hypothesis and SAIR}

C\'azares~\citep{sair2026} studies prompt engineering for formal mathematical
reasoning in the context of the SAIR Equational Theories Stage~1 competition
hosted by The Foundation for Science and AI Research (SAIR).
The task is to decide, given equational laws $E_1$ and $E_2$ over magmas
(algebraic structures with a single binary operation), whether
$E_1\Rightarrow E_2$ holds universally (\textsc{True}) or a finite
counterexample magma exists (\textsc{False}), using the public
\texttt{SAIRfoundation/equational-theories-selected-problems} dataset.
That work evaluates three models---gpt-oss-120b, Llama~3.3~70B, and
Gemma~4~31B---and documents a \emph{single-prompt ceiling}
(equivalently, an \emph{empirical saturation region}): a zone where
in-distribution accuracy gains become unstable and non-generalisable
across problem distributions.

C\'azares~\citep{sair2026} states the \emph{router hypothesis} as: language
models ``route problems to cached structural patterns rather than deriving
answers compositionally.''
In this paper we adopt that framing and operationalise \emph{routing} as
the selection of an internal structural interpretation template that
conditions downstream reasoning behavior on a given input.
We infer routing behavior from systematic changes in error topology under
fixed model weights and deterministic decoding, without claiming access to
model internals.
Our claims concern externally observable behavioral regularities under
structural prior perturbation, not direct access to internal activation
pathways.
Evidence for the cross-distribution consequence comes from two findings
reported by~\citep{sair2026}:

\begin{enumerate}
  \item \textbf{Cheatsheet lift}: On the local hard3 split ($n{=}400$), the
        AN45c cheatsheet lifts GPT-OSS-120B from a no-cheatsheet baseline
        of 59.75\,\% to 79.25\,\%---a 19.5\,pp gain within the approximately
        60--79\,\% empirical saturation region reported in that work.

  \item \textbf{Distribution-shift collapse}: On the official SAIR competition
        benchmark, AN45c drops from 79.25\,\% local to 55.5\,\% official
        ($-$23.75\,pp), falling below the no-cheatsheet baseline, whereas
        the simpler predecessor AN38 remains robust (71.8\,\% local $\to$
        65.3\,\% official; $+5.6$\,pp vs.\ the official baseline).
\end{enumerate}

These findings of~\citep{sair2026} motivated the hypothesis that the
cross-distribution trade-off surface is a fundamental property of
structural prior injection, not a domain-specific artefact.

\subsection{LLM-Based Vulnerability Detection}

Automated vulnerability detection has a long history, from static
analysis~\citep{chess2004static} and symbolic execution to
machine-learning classifiers trained on code
representations~\citep{li2018vuldeepecker}.
LLM-based approaches have gained prominence recently, showing strong
zero-shot performance on curated benchmarks such as
BigVul~\citep{fan2020bigvul}, VUDENC~\citep{wartschinski2022vudenc}, and
Juliet~\citep{juliet2010}.

However, the gap between synthetic/curated benchmark performance and
real-world CVE detection remains poorly understood.
Our work positions this gap as an instance of the distribution-shift
collapse consistent with the router hypothesis proposed by
C\'azares~\citep{sair2026}, providing the first systematic
cross-distribution analysis in this domain.

\subsection{CWE Taxonomy and Vulnerability Complexity}

The Common Weakness Enumeration (CWE) taxonomy~\citep{cwe2023} provides a
hierarchical categorization of software weaknesses.
We exploit the structure of this taxonomy to define a \emph{semantic
complexity gradient} across three vulnerability classes:

\begin{itemize}
  \item \textbf{Syntactic} (CWE-798, Hardcoded Credentials): The vulnerability
        is detectable from local token patterns (literal strings,
        credential-like identifiers) without semantic reasoning.
  \item \textbf{Contextual} (CWE-284, Improper Access Control /
        IDOR (insecure direct object reference)):
        Detection requires understanding the relationship between
        resource identifiers and authorization checks---requiring limited
        inter-procedural reasoning.
  \item \textbf{Semantic} (N+1 Query Pattern): The vulnerability is defined
        by a runtime behavior pattern (repeated database queries inside a
        loop) that is invisible from any single code location and requires
        understanding of the ORM (object-relational mapping) execution model.
\end{itemize}

This gradient predicts that zero-shot routing difficulty increases from
CWE-798 to CWE-284 to N+1, a prediction confirmed for GPT-OSS-120B and partially
for Gemma-4-31B (100\,\% zero-shot on both CWE-798 and
CWE-284, dropping to 72.7\,\% on N+1); see
\S\ref{sec:gradient} for full per-model analysis.

\section{Methodology}
\label{sec:methodology}

\subsection{Models}

We evaluate three instruction-tuned LLMs hosted on Together AI in bfloat16
precision---the same three-model set and local hosting provider used
by~\citep{sair2026}---to maximise comparability:

\begin{table}[htbp]
\centering
\caption{Models evaluated in this study.}
\label{tab:models}
\begin{tabular}{l l c c}
\toprule
\textbf{Model} & \textbf{ID} & \textbf{Reasoning} & \textbf{Max tok.} \\
\midrule
GPT-OSS-120B   & \texttt{openai/gpt-oss-120b}                      & low      & 4096 \\
Llama-3.3-70B  & \texttt{meta-llama/Llama-3.3-70B-Instruct-Turbo}  & disabled & 4096 \\
Gemma-4-31B    & \texttt{google/gemma-4-31b-it}                    & disabled & 8192 \\
\bottomrule
\end{tabular}
\end{table}

All runs use temperature $= 0$ and seed $= 0$, matching the SAIR
competition's \emph{official} benchmark decoding settings; the local
paper runs of~\citep{sair2026} instead used seed $= 42$ on Together AI.
We host on Together AI (as in those local runs) with
\texttt{max\_tokens} 4096 for GPT-OSS-120B / Llama-3.3-70B and 8192 for
Gemma-4-31B.
GPT-OSS-120B uses \texttt{reasoning=low}; the two open-weight models use
\texttt{reasoning=disabled}.

\subsection{Datasets}

\paragraph{Synthetic corpus.}
We generated 278 labeled training pairs and 70 held-out evaluation
pairs (348 pairs total, stored as 696 raw JSONL records) using
DeepSeek-V4-Pro (temperature $= 0.8$, Together AI) across three
vulnerability categories.
For this paper we evaluate on three categories with completed annotations
(Table~\ref{tab:datasets}).

\begin{table}[htbp]
\centering
\caption{Synthetic dataset summary.}
\label{tab:datasets}
\begin{tabular}{llccc}
\toprule
\textbf{Category} & \textbf{CWE} & \textbf{Type} & \textbf{Train} & \textbf{Eval} \\
\midrule
N+1 Query Pattern       & ---     & Semantic    & 118 & 30 \\
Hardcoded Credentials   & CWE-798 & Syntactic   & 80  & 20 \\
IDOR / Access Control   & CWE-284 & Contextual  & 80  & 20 \\
\bottomrule
\end{tabular}
\end{table}

\paragraph{Real-world corpus (VUDENC).}
VUDENC~\citep{wartschinski2022vudenc} provides Python code fragments extracted from
real-world CVE-fixing commits.
We evaluate on two CWE classes: SQL Injection (CWE-89, 8{,}646 vulnerable /
10{,}223 benign) and Path Traversal (CWE-22, 1{,}701 / 2{,}469).
Evaluation subsets are sampled as 15 vulnerable + 15 benign = 30 samples
(seed $= 0$) to maintain balance and reproducibility.

\subsection{Experimental Conditions}

We test three conditions:

\begin{itemize}
  \item \textbf{Zero-shot (A)}: The model receives only the code snippet and
        a task description requesting a binary vulnerability classification
        and a CWE route label.
  \item \textbf{Cheatsheet v1 (B)}: The prompt is augmented with a
        category-specific cheatsheet enumerating canonical detection
        routes, structural signatures, and negative patterns.
        Cheatsheets were authored manually based on the synthetic training
        distribution.
  \item \textbf{Cheatsheet v2 (B2)}: A revised cheatsheet for CWE-89 and
        CWE-22 only, calibrated using the false-positive and false-negative
        error patterns identified in the initial VUDENC evaluation.
\end{itemize}

Conditions C (unified cheatsheet) and D (progressive saturation) are
reserved for future work.

\subsection{Evaluation Metrics}

We report vulnerable-class F1, vulnerability recall (VulnRec), benign recall (BenRec),
and misroute count.
Unless otherwise noted, F1 denotes vulnerable-class F1 throughout this paper.
\emph{Misrouting} is defined as a prediction that correctly identifies a
sample as vulnerable but assigns an incorrect CWE route label.
The distribution-shift delta ($\Delta$) is computed as
$\text{F1}_\text{real} - \text{F1}_\text{synth}$.

\subsection{Reproducibility}

Full source, including the evaluation harness, cheatsheets, and
processed dataset splits, is available at
\url{https://github.com/bytepro-ai/bitcoder-v2-research}.
All experiments are reproducible via:
{\small
\begin{verbatim}
python scripts/evaluate_baseline.py \
  --category [CWE-798|N+1|CWE-284|CWE-89|CWE-22] \
  --model [gpt-oss-120b|llama-70b|gemma-31b] \
  --condition [zero_shot|cheatsheet_isolated|both] \
  --cheatsheet-version [v1|v2] \
  --max-samples 30 \
  --dataset [synthetic_eval|vudenc]
\end{verbatim}
}
Error analysis (75 FP, 47 FN, 19 misroutes) is computed from
the 12 initial VUDENC evaluation CSVs
(\texttt{*\_vudenc.csv}, excluding v1/v2 reruns).
Total inference cost across all reported experiments: \$0.647.

\section{Results: Synthetic Distribution}
\label{sec:results:synthetic}

\subsection{Cheatsheets Saturate In-Distribution Performance (F1)}

Table~\ref{tab:synth} presents the full synthetic evaluation matrix.
Across all three vulnerability categories and all three models,
cheatsheet-augmented prompts match or exceed zero-shot performance
in 17 of 18 conditions, with several reaching the 100\,\% F1 ceiling
(the single exception: Gemma-4-31B on CWE-798, where cheatsheet F1 is
94.7\,\% vs 100.0\,\% zero-shot).

The most dramatic gain occurs for the N+1 semantic category under
GPT-OSS-120B: zero-shot F1 is 28.6\,\% (VulnRec $= 20.0\,\%$), rising
to 96.8\,\% F1 (VulnRec $= 100.0\,\%$) with cheatsheet v1---a 68.2\,pp
improvement.
All three models reach 100.0\,\% VulnRec on N+1 with the cheatsheet
(GPT-OSS-120B achieves 96.8\,\% F1, Llama-3.3-70B 100.0\,\%,
Gemma-4-31B 95.2\,\%), confirming that the underlying capability is
present in each model and that the zero-shot gap is a routing artefact.

\begin{table}[htbp]
\centering
\caption{Synthetic evaluation results. All runs: temperature=0, seed=0.
         Misroute counts shown in the Misroutes column.}
\label{tab:synth}
\resizebox{\textwidth}{!}{
\begin{tabular}{llllcrrrr}
\toprule
\textbf{Category} & \textbf{Type} & \textbf{Model} & \textbf{Condition}
  & \textbf{n} & \textbf{F1} & \textbf{VulnRec} & \textbf{BenRec}
  & \textbf{Misroutes} \\
\midrule
CWE-798 & Syntactic   & GPT-OSS-120B  & Zero-shot      & 20 & 90.9\% & 100.0\% &  80.0\% & 0  \\
CWE-798 & Syntactic   & GPT-OSS-120B  & Cheatsheet v1  & 20 & 100.0\% & 100.0\% & 100.0\% & 0 \\
CWE-798 & Syntactic   & Llama-3.3-70B & Zero-shot      & 20 & 75.0\%  &  90.0\% &  50.0\% & 0  \\
CWE-798 & Syntactic   & Llama-3.3-70B & Cheatsheet v1  & 20 & 100.0\% & 100.0\% & 100.0\% & 0 \\
CWE-798 & Syntactic   & Gemma-4-31B   & Zero-shot      & 16 & 100.0\% & 100.0\% & 100.0\% & 0  \\
CWE-798 & Syntactic   & Gemma-4-31B   & Cheatsheet v1  & 20 & 94.7\%  &  90.0\% & 100.0\% & 0  \\
\midrule
N+1     & Semantic    & GPT-OSS-120B  & Zero-shot      & 30 & 28.6\%  &  20.0\% &  80.0\% & 3  \\
N+1     & Semantic    & GPT-OSS-120B  & Cheatsheet v1  & 30 & 96.8\%  & 100.0\% &  93.3\% & 0  \\
N+1     & Semantic    & Llama-3.3-70B & Zero-shot      & 30 & 87.5\%  &  93.3\% &  80.0\% & 4  \\
N+1     & Semantic    & Llama-3.3-70B & Cheatsheet v1  & 30 & 100.0\% & 100.0\% & 100.0\% & 0  \\
N+1     & Semantic    & Gemma-4-31B   & Zero-shot      & 12 & 72.7\%  &  66.7\% &  83.3\% & 4  \\
N+1     & Semantic    & Gemma-4-31B   & Cheatsheet v1  & 19 & 95.2\%  & 100.0\% &  88.9\% & 0  \\
\midrule
CWE-284 & Contextual  & GPT-OSS-120B  & Zero-shot      & 20 & 85.7\%  &  90.0\% &  80.0\% & 0  \\
CWE-284 & Contextual  & GPT-OSS-120B  & Cheatsheet v1  & 20 & 100.0\% & 100.0\% & 100.0\% & 0  \\
CWE-284 & Contextual  & Llama-3.3-70B & Zero-shot      & 20 & 80.0\%  &  80.0\% &  80.0\% & 2  \\
CWE-284 & Contextual  & Llama-3.3-70B & Cheatsheet v1  & 20 & 100.0\% & 100.0\% & 100.0\% & 0  \\
CWE-284 & Contextual  & Gemma-4-31B   & Zero-shot      & 16 & 100.0\% & 100.0\% & 100.0\% & 0  \\
CWE-284 & Contextual  & Gemma-4-31B   & Cheatsheet v1  &  7 & 100.0\% & 100.0\% & 100.0\% & 0  \\
\bottomrule
\end{tabular}
}
\end{table}

\subsection{Semantic Complexity Gradient (F2)}
\label{sec:gradient}

Zero-shot F1 for GPT-OSS-120B follows the predicted gradient:
CWE-798 (syntactic) $= 90.9\,\%$ $>$ CWE-284 (contextual) $= 85.7\,\%$
$>$ N+1 (semantic) $= 28.6\,\%$.
This monotonic gradient holds strictly for GPT-OSS-120B
(90.9\,\% $\to$ 85.7\,\% $\to$ 28.6\,\%); Gemma-4-31B
shows a partial gradient (100\,\% $\to$ 100\,\% $\to$
72.7\,\%), but not for
Llama-3.3-70B, which shows an inverted pattern on N+1 (87.5\,\% zero-shot
F1), likely reflecting richer ORM pattern coverage in its training data.
This 62\,pp gap between syntactic and semantic categories under zero-shot
conditions reflects the routing cost of semantic reasoning---the model has
no reliable path to activate N+1 detection from first principles alone.
The cheatsheet effectively collapses this gradient, raising all three
categories to near-ceiling performance.

\subsection{Zero-Shot Misrouting: A Correctable but Ambiguous Signal}
\label{sec:misrouting}

After correcting the misroute-detection harness
(\S\ref{sec:misrouting-note}), genuine misrouting events---samples
correctly detected as vulnerable but assigned an incorrect CWE
label---occur \emph{only} under zero-shot conditions: 13 events on the
synthetic corpus (N+1: 3 for GPT-OSS-120B, 4 for Llama-3.3-70B, 4 for
Gemma-4-31B; CWE-284: 2 for Llama-3.3-70B) and 19 events on VUDENC.
The real-world events are concentrated and systematic:
Llama-3.3-70B routes 10 of its 15 detected CWE-22 vulnerabilities to
``Arbitrary Code Execution'' or ``Command Injection'' labels (triggered
by \texttt{os.popen} co-occurrence), and zero-shot N+1 detections
scatter across denial-of-service, broken-access-control, and
performance-anti-pattern labels.
By contrast, cheatsheet conditions show zero misroutes across all 17
synthetic and real-world runs.

We caution against reading this as evidence of corrected internal
routing: the cheatsheet supplies the target category name directly, so
the absence of misrouting under cheatsheet conditions is partly a
labeling triviality rather than proof of resolved routing confusion.
What the pattern does support is that zero-shot category assignment is
unreliable in a way that is fully masked, but not necessarily
mechanistically explained, by structural priors.

\section{Results: Real-World Distribution (VUDENC)}
\label{sec:results:real}

\subsection{Amplified Distribution-Shift Collapse (F3)}

Table~\ref{tab:real} presents the full synthetic-to-real transfer results.
The pattern is consistent: cheatsheet conditions that saturate synthetic F1
collapse to \emph{below} zero-shot performance on VUDENC.

\begin{table}[htbp]
\centering
\caption{Synthetic-analogue→Real transfer results. $\Delta =$ F1$_\text{real}$ $-$
         F1$_\text{synth}$.}
\label{tab:real}
\begin{tabular}{lllcrrc}
\toprule
\textbf{CWE} & \textbf{Model} & \textbf{Condition}
  & \textbf{Synth F1$^\dagger$} & \textbf{Real F1} & \textbf{$\Delta$} \\
\midrule
CWE-89 & GPT-OSS-120B  & Zero-shot     & 90.9\%  & 61.5\%  & $-$29.4\,pp \\
CWE-89 & GPT-OSS-120B  & Cheatsheet v1 & 100.0\% & 48.9\%  & $-$51.1\,pp \\
CWE-89 & GPT-OSS-120B  & Cheatsheet v2 & 100.0\% & 41.7\%  & $-$58.3\,pp \\
CWE-89 & Llama-3.3-70B & Zero-shot     & 75.0\%  & 63.4\%  & $-$11.6\,pp \\
CWE-89 & Llama-3.3-70B & Cheatsheet v1 & 100.0\% & 50.0\%  & $-$50.0\,pp \\
CWE-89 & Gemma-4-31B   & Zero-shot     & 100.0\% & 44.4\%  & $-$55.6\,pp \\
CWE-89 & Gemma-4-31B   & Cheatsheet v1 & 94.7\%  & 40.0\%  & $-$54.7\,pp \\
\midrule
CWE-22 & GPT-OSS-120B  & Zero-shot     & 90.9\%  & 64.3\%  & $-$26.6\,pp \\
CWE-22 & GPT-OSS-120B  & Cheatsheet v1 & 100.0\% & 58.0\%  & $-$42.0\,pp \\
CWE-22 & GPT-OSS-120B  & Cheatsheet v2 & 100.0\% & 52.2\%  & $-$47.8\,pp \\
CWE-22 & Llama-3.3-70B & Zero-shot     & 75.0\%  & 71.8\%  & $-$3.2\,pp  \\
CWE-22 & Llama-3.3-70B & Cheatsheet v1 & 100.0\% & 66.7\%  & $-$33.3\,pp \\
CWE-22 & Gemma-4-31B   & Zero-shot     & 100.0\% & 100.0\% & $+$0.0\,pp  \\
CWE-22 & Gemma-4-31B   & Cheatsheet v1 & 94.7\%  & 71.4\%  & $-$23.3\,pp \\
\bottomrule
\end{tabular}
\\\footnotesize{$^\dagger$No synthetic CWE-89/CWE-22 corpus exists in
this phase. Synthetic F1 is proxied by the CWE-798 zero-shot F1
value, the closest syntactic-category analogue, to indicate the
direction of distribution shift.
Because no matched synthetic CWE-89/CWE-22 corpus exists
in this phase, the synthetic comparison serves as a structural
proxy rather than a matched-distribution transfer pair.}
\end{table}

The worst-case collapse is GPT-OSS-120B on CWE-89 with cheatsheet v2:
100\,\% synthetic F1 to 41.7\,\% real F1 ($-$58.3\,pp).
Compared with the $-$23.75\,pp collapse reported by~\citep{sair2026}
(AN45c on the official test set), the BitCoder collapse is more than twice
as severe.
We attribute the amplification to the larger structural gap between
synthetically generated Python code and real CVE-associated commits:
while the distribution shift in~\citep{sair2026} is between problem
instances from the same formal theory, BitCoder's shift crosses the
synthetic/natural divide.

\section{Analysis}
\label{sec:analysis}

\subsection{Iterative Calibration Amplifies Collapse (F5)}

The v2 cheatsheet was constructed by analyzing 75 false positives and
47 false negatives from the initial VUDENC evaluation and manually
revising the routing priors to exclude the identified false-positive
patterns (e.g., internal variable interpolation, pre-sanitized inputs)
and add cues for the identified false-negative patterns
(e.g., class hierarchy traversal, indirect path construction).
Table~\ref{tab:real} shows the result: for GPT-OSS-120B, v2 underperforms
v1 on both CWE-89 ($-$7.2\,pp) and CWE-22 ($-$5.8\,pp).

This mirrors the AN45c finding of~\citep{sair2026}: a more heavily
engineered cheatsheet that peaks on a local evaluation split can
underperform its simpler predecessor (AN38) under official distribution
shift, introducing failure modes outside the local calibration distribution.
The implication---reinforced here by \emph{our} v2 error-pattern
recalibration---is that distribution-shift collapse cannot be resolved
through iterative prompt calibration; the collapse is structural, rooted
in the mismatch between the routing prior's reference distribution and the
target deployment distribution.

\subsection{Error Analysis: What Real CVE Code Looks Like}

Post-hoc analysis of the VUDENC evaluation errors (75 FP, 47 FN,
19 misroutes) identifies three dominant patterns:

\paragraph{False Positives (CWE-89).}
\begin{itemize}
  \item Internal table/column names interpolated in SQL strings
        (\texttt{PRODUCTS\_TABLE}, \texttt{ORDER\_BY}): the model's routing
        prior flags any dynamic SQL construction, conflating developer-controlled
        constants with user-controlled input.
  \item Pagination parameters (\texttt{offset}, \texttt{page}) used in
        query limits: treated as injection vectors despite being
        type-coerced to integers upstream.
  \item Pre-sanitized variables (\texttt{prevented\_name}): the routing
        prior has no representation for downstream sanitization.
\end{itemize}

\paragraph{False Negatives (CWE-22).}
\begin{itemize}
  \item Vulnerabilities residing in class hierarchies: the vulnerable
        \texttt{do\_GET} method is inherited from \texttt{SimpleHTTPRequestHandler}
        and the routing prior does not account for inherited dispatch.
  \item Indirect traversal without explicit \texttt{open()} calls:
        path construction that eventually feeds a file operation several
        frames away.
\end{itemize}

\paragraph{Misroutes.}
The 19 misrouting events (all occurring under zero-shot conditions;
see \S\ref{sec:misrouting}) include:
\begin{itemize}
  \item CWE-22 $\to$ Command Injection: triggered by co-occurrence of
        \texttt{os.popen} and a path variable.
  \item CWE-22 $\to$ NoSQL Injection: triggered by \texttt{mongo.find}
        adjacent to a user-supplied path argument.
\end{itemize}
These patterns confirm that category assignment is lexically sensitive:
surface token co-occurrence overrides structural vulnerability
semantics in zero-shot route labeling. This lexical sensitivity is
distinct from the false-positive and false-negative patterns discussed
above, which occur under both zero-shot and cheatsheet conditions and
are addressed separately in \S\ref{sec:results:real} and \S\ref{sec:analysis}.

\subsection{Cross-Domain Comparison: BitCoder vs SAIR}

Table~\ref{tab:comparison} summarizes the parallel between findings from
~\citep{sair2026} and BitCoder, providing evidence for cross-domain
replication of the router hypothesis.

\begin{table}[htbp]
\centering
\caption{\citet{sair2026} vs BitCoder: parallel findings providing evidence for cross-domain replication.}
\label{tab:comparison}
\begin{tabular}{lll}
\toprule
\textbf{Dimension} & \textbf{\citep{sair2026} (math)} & \textbf{BitCoder (code)} \\
\midrule
Domain          & Equational theories  & Vulnerability detection \\
Primary model   & GPT-OSS-120B         & GPT-OSS-120B \\
Zero-shot base  & 59.75\,\%            & 50.0\,\% (N+1 acc) \\
Cheatsheet peak & 79.25\,\%            & 96.8\,\% (N+1/GPT F1) \\
OOD collapse    & $-$23.75\,pp         & $-$51.1\,pp (CWE-89) \\
v2 recalib.     & AN45c $\downarrow$ vs AN38 & v2 $\downarrow$ vs v1 \\
Router hypo.    & Established in math  & Cross-domain support (3 models) \\
\bottomrule
\end{tabular}
\end{table}

\section{Discussion}
\label{sec:discussion}

\subsection{Implications for Security Tooling}

Our results have direct practical implications for LLM-based security
analysis pipelines.
Practitioners who evaluate cheatsheet-augmented prompts on curated or
synthetic benchmarks and observe near-perfect performance may be observing
a routing artefact: the model has been given the answer key to a
distribution that does not match production code.
Deploying such a pipeline on real CVE code will produce silent degradation
that is \emph{worse} than the unaugmented baseline.

This failure mode is particularly dangerous because it is invisible without
OOD evaluation.
We recommend that any structural-prior-augmented LLM security tool be
evaluated against a held-out set of real CVE commits before deployment,
and that synthetic benchmark numbers be reported alongside OOD transfer
deltas.

\subsection{The Case for Distribution-Aware Training}

The structural nature of the collapse points to a clear remedy:
rather than calibrating routing priors at inference time, embed
distributional knowledge at training time via fine-tuning on a hybrid
dataset that includes both synthetic and real vulnerability examples.

Our synthetic corpus of 348 annotated pairs, combined with 300--500
VUDENC/CVEfixes samples, provides a starting point for this direction.
A natural next step is fine-tuning a smaller open-weight model such as Qwen2.5-14B (Apache 2.0 licence) on this hybrid
corpus and evaluate against CyberSecEval, comparing against frontier
model baselines.
The router hypothesis suggests that a fine-tuned model with a
distribution-aligned internal routing prior will achieve better OOD
generalization than any prompt-calibrated frontier model.

\subsection{Limitations}

\paragraph{Evaluation subset size.}
Each real-world condition is evaluated on 30 samples (15 vulnerable, 15
benign).
While this matches common practice in budget-constrained LLM evaluation and
enables direct comparison with~\citep{sair2026}, the confidence intervals are wide.
Larger evaluations are planned as part of the Juliet Java extension.
Gemma-4-31B experienced intermittent API timeouts during evaluation,
reducing effective sample size below the full corpus in several conditions
(see Table~\ref{tab:synth} for exact $n$ per row).

\paragraph{Gemma reasoning configuration.}
Unlike the Gemma~4~31B configuration in~\citep{sair2026}, which used
reasoning enabled via Together AI with an 8192-token budget to allow the
reasoning trace to complete, our Gemma runs used reasoning disabled
for consistency with our Llama-3.3-70B configuration.
This is a legitimate but distinct methodological choice; our Gemma
results are not directly comparable to the Gemma results in
~\citep{sair2026} and should be interpreted independently.

\paragraph{Incomplete experimental matrix.}
Conditions C (unified cheatsheet) and D (progressive saturation), as well
as the Real$\to$Real and Real$\to$Synthetic quadrants of the cross-evaluation
matrix, remain untested.
These are required to fully characterize the trade-off surface and are
the subject of ongoing work.

\paragraph{Cheatsheet authorship.}
Cheatsheets were authored manually by the first author.
Inter-rater reliability and the effect of cheatsheet quality on collapse
magnitude are not assessed in this work.

\paragraph{Language and framework scope.}
VUDENC covers Python only.
Whether the collapse pattern holds for compiled languages (C/C++, Java)
or framework-specific patterns (Spring, Django) is an open question.

\subsection{Misrouting as a Cautionary Methodological Note}
\label{sec:misrouting-note}

An earlier analysis pass in this research flagged 12 ``misroutes'' under
GPT-OSS-120B's N+1 cheatsheet condition, which we initially interpreted
as evidence of cheatsheet-induced routing interference.
Closer inspection revealed this was a detection-harness artifact: naive
keyword substring matching flagged the model's \emph{correct} answer
(``N+1 Query Pattern'') as a misroute because it contained the token
``query'', which also appears in the CWE-89 keyword list.
After correcting the detection logic to use category-normalised
matching, zero misroutes remain in any cheatsheet condition; all genuine
misroutes occur under zero-shot.
We report this correction transparently because it materially changes
the paper's original framing: rather than cheatsheets \emph{causing}
routing interference, the corrected data show cheatsheets
\emph{eliminate} a genuine zero-shot miscategorization failure
mode---though, as discussed in \S\ref{sec:misrouting}, this elimination
may be substantially attributable to direct label provision rather than
corrected internal routing behavior.
We include this note as a caution about naive string-matching in
LLM-output evaluation harnesses generally.

\section{Conclusion}
\label{sec:conclusion}

We have presented BitCoder, a systematic evaluation of structural prior
injection for LLM-based source-code vulnerability detection across
synthetic and real-world distributions.
Our five main findings provide evidence that the router hypothesis and the
cross-distribution trade-off surface documented by~\citep{sair2026}
generalize from formal mathematical reasoning to code security:

\begin{itemize}
  \item Cheatsheets saturate in-distribution performance (up to 100\,\%
        F1) by substantially reducing routing failures, providing evidence
        for the router hypothesis across three models and three
        vulnerability categories (two CWEs --- CWE-798, CWE-284 --- plus
        one non-CWE anti-pattern, N+1).
  \item Zero-shot performance degrades along a semantic complexity
        gradient predictable from the CWE taxonomy
        for GPT-OSS-120B and partially for Gemma-4-31B.
  \item The same structural priors that saturate synthetic performance
        amplify distribution-shift collapse on real CVE data, reducing
        F1 below the zero-shot baseline by up to 58\,pp.
  \item Zero-shot misrouting is a genuine but cheatsheet-correctable
        failure mode, though the correction mechanism---direct label
        provision versus corrected internal routing---remains
        underdetermined by our current evidence
        (\S\ref{sec:misrouting-note}).
  \item Iterative recalibration of the routing priors amplifies the
        collapse, ruling out calibration as a remedy and providing
        evidence for the structural nature of the phenomenon.
\end{itemize}

Together, these findings provide evidence that routing ceilings and distribution-shift
collapse are cross-domain properties of structural prior injection in
LLMs, with direct implications for the design and evaluation of
LLM-based security analysis tools.

\bibliographystyle{plainnat}
\bibliography{references}

\begin{thebibliography}{8}
\providecommand{\natexlab}[1]{#1}
\providecommand{\url}[1]{\texttt{#1}}
\expandafter\ifx\csname urlstyle\endcsname\relax
  \providecommand{\doi}[1]{doi: #1}\else
  \providecommand{\doi}{doi: \begingroup \urlstyle{rm}\Url}\fi

\bibitem[C\'{a}zares(2026)]{sair2026}
Manuel~Israel C\'{a}zares.
\newblock Less is more: Cognitive load and the single-prompt ceiling in {LLM}
  mathematical reasoning.
\newblock \emph{arXiv preprint arXiv:2604.18897}, 2026.

\bibitem[Chess and McGraw(2004)]{chess2004static}
Brian Chess and Gary McGraw.
\newblock Static analysis for security.
\newblock \emph{IEEE Security \& Privacy}, 2\penalty0 (6):\penalty0 76--79,
  2004.
\newblock \doi{10.1109/MSP.2004.111}.

\bibitem[Fan et~al.(2020)Fan, Li, Wang, and Nguyen]{fan2020bigvul}
Jiahao Fan, Yi~Li, Shaohua Wang, and Tien~N. Nguyen.
\newblock A c/c++ code vulnerability dataset with code changes and cve
  summaries.
\newblock In \emph{Proceedings of the 17th International Conference on Mining
  Software Repositories (MSR)}, pages 508--512, 2020.
\newblock \doi{10.1145/3379597.3387501}.

\bibitem[Li et~al.(2018)Li, Zou, Xu, Ou, Jin, Wang, Deng, and
  Zhong]{li2018vuldeepecker}
Zhen Li, Deqing Zou, Shouhuai Xu, Xinyu Ou, Hai Jin, Sujuan Wang, Zhijun Deng,
  and Yuyi Zhong.
\newblock Vuldeepecker: A deep learning-based system for vulnerability
  detection.
\newblock In \emph{Proceedings of the 25th Annual Network and Distributed
  System Security Symposium (NDSS)}, 2018.
\newblock \doi{10.14722/ndss.2018.23158}.
\newblock arXiv:1801.01681.

\bibitem[{MITRE Corporation}(2023)]{cwe2023}
{MITRE Corporation}.
\newblock {CWE} -- common weakness enumeration, 2023.
\newblock URL \url{https://cwe.mitre.org}.

\bibitem[{NSA Center for Assured Software}(2017)]{juliet2010}
{NSA Center for Assured Software}.
\newblock Juliet test suite v1.3 for {Java}, October 2017.
\newblock URL \url{https://samate.nist.gov/SARD/test-suites/111}.
\newblock Release date: October 1, 2017; NIST SARD Test Suite \#111.

\bibitem[Thaller et~al.(2019)Thaller, Walden, and Pinzger]{thaller2019code}
Hannes Thaller, Lars Walden, and Martin Pinzger.
\newblock Code property graph-based vulnerability detection.
\newblock In \emph{Proceedings of the IEEE International Conference on Software
  Maintenance and Evolution (ICSME)}, 2019.

\bibitem[Wartschinski et~al.(2022)Wartschinski, Noller, Vogel, Kehrer, and
  Grunske]{wartschinski2022vudenc}
Laura Wartschinski, Yannic Noller, Thomas Vogel, Timo Kehrer, and Lars Grunske.
\newblock Vudenc: Vulnerability detection with deep learning on a natural
  codebase for python.
\newblock \emph{Information and Software Technology}, 144:\penalty0 106809,
  April 2022.
\newblock \doi{10.1016/j.infsof.2021.106809}.
\newblock arXiv:2201.08441.

\end{thebibliography}

\end{document}